\documentclass[11pt,a4paper]{article}
\usepackage[hyphens]{url}
\usepackage[hyperref]{acl2017}
\usepackage{times}
\usepackage{latexsym}
\usepackage[hyphenbreaks]{breakurl}
\usepackage{graphicx}
\usepackage[caption=false,font=footnotesize]{subfig}
\usepackage{amsmath}
\usepackage{fancyhdr} 
\aclfinalcopy 

\title{Using Global Constraints and Reranking to Improve Cognates Detection}

\author{Michael Bloodgood \\
  Department of Computer Science \\
  The College of New Jersey \\
  Ewing, NJ 08628 \\
  {\tt mbloodgood@tcnj.edu} \\\And
  Benjamin Strauss \\
  Computer Science and Engineering Dept. \\
  The Ohio State University \\
  Columbus, OH 43210 \\
  {\tt strauss.105@osu.edu} \\}

\date{}

\begin{document}

\thispagestyle{fancy}

\maketitle

\begin{abstract}
Global constraints and reranking have not been used in cognates detection research to date. We propose methods for using global constraints by performing rescoring of the score matrices produced by state of the art cognates detection systems.
Using global constraints to perform rescoring is complementary to state of the art methods for performing cognates detection and results in significant performance improvements beyond current state of the art performance on publicly available datasets with different language pairs and various conditions such as different levels of baseline state of the art performance and different data size conditions, including with more realistic large data size conditions than have been evaluated with in the past. 
\end{abstract}

\section{Introduction} \label{introduction}

This paper presents an effective method for using global constraints to improve performance for cognates detection. 
Cognates detection is the task of identifying words across languages that have a common origin. 
Automatic cognates detection is important to linguists because cognates are needed to determine how languages evolved. 
Cognates are used for protolanguage reconstruction \cite{hall2011,bouchard2013}. 
Cognates are important for cross-language dictionary look-up and can also improve the quality of machine translation, word alignment, and bilingual lexicon induction \cite{simard1993,kondrak2003}.

A word is traditionally only considered cognate with another if both words proceed from the same ancestor.  Nonetheless, in line with the
conventions of previous research in computational linguistics,  we set a broader definition.   We use the word `cognate' to denote, as in \cite{kondrak2001}:  ``...words in different languages that are similar in form and meaning, without
making a distinction between borrowed and genetically related words; for example, English `sprint' and the Japanese
borrowing `supurinto' are considered cognate, even though
these two languages are unrelated."  These broader criteria
are motivated by the ways scientists develop and use cognate identification algorithms in natural language processing (NLP) systems.  
For cross-lingual applications, the advantage of such technology is the ability to identify words
for which similarity in meaning can be accurately inferred
from similarity in form; it does not matter if the similarity
in form is from strict genetic relationship or later borrowing \cite{mericli2012}.

Cognates detection has received a lot of attention in the literature. The research of the use of statistical learning methods to build systems that can automatically perform cognates detection has yielded many interesting and creative approaches for gaining traction on this challenging task. Currently, the highest-performing state of the art systems detect cognates based on the combination of multiple sources of information. Some of the most indicative sources of information discovered to date are word context information, phonetic information, word frequency information, temporal information in the form of word frequency distributions across parallel time periods, and word burstiness information. See section~\ref{experiments} for fuller explanations of each of these sources of information that state of the art systems currently use. Scores for all pairs of words from language L1 x language L2 are generated by generating component scores based on these sources of information and then combining them in an appropriate manner. Simple methods of combination are giving equal weighting for each score, while state of the art performance is obtained by learning an optimal set of weights from a small seed set of known cognates. 
Once the full matrix of scores is generated, the word pairs with the highest scores are predicted as being cognates.

The methods we propose in the current paper consume as input the final score matrix that state of the art methods create.  We test if our methods can improve performance by generating new rescored matrices by rescoring all of the pairs of words by taking into account global constraints that apply to cognates detection. Thus, our methods are complementary to previous methods for creating cognates detection systems. Using global constraints and performing rescoring to improve cognates detection has not been explored yet. We find that rescoring based on global constraints improves performance significantly beyond current state of the art levels. 

The cognates detection task is an interesting task to apply our methods to for a few reasons:
\begin{itemize}
\item It's a challenging unsolved task where ongoing research is frequently reported in the literature trying to improve performance;
\item There is significant room for improvement in performance;
\item It has a global structure in its output classifications since if a word lemma\footnote{A lemma is a base form of a word. For example, in English the words `baked' and `baking' would both map to the lemma `bake'. Lemmatizing software exists for many languages and lemmatization is a standard preprocessing task conducted before cognates detection.} $w_i$ from language L1 is cognate with a word lemma $w_j$ from language L2, then $w_i$ is not cognate with any other word lemma from L2 different from $w_j$ and $w_j$ is not cognate with any other word lemma $w_k$ from L1. 
\item There are multiple standard datasets freely and publicly available that have been worked on with which to compare results. 
\item Different datasets and language pairs yield initial score matrices with very different qualities. Some of the score matrices built using the existing state of the art best approaches yield performance that is quite low (11-point interpolated average precision of only approximately 16\%) while some of these score matrices for other language pairs and data sets have state of the art score matrices that are already able to achieve 11-point interpolated average precision of 57\%.
\end{itemize}

Although we are not aware of work using global constraints to perform rescoring to improve cognates detection, there are related methodologies for reranking in different settings. Methodologically related work includes past work in structured prediction and reranking \cite{collins2002,collins2004,collins2005,taskar2005a,taskar2005b}. 
Note that in these past works, there are many instances with structured outputs that can be used as training data to learn a structured prediction model. For example, a seminal application in the past was using online training with structured perceptrons to learn improved systems for performing various syntactic analyses and tagging of sentences such as POS tagging and base noun phrase chunking \cite{collins2002}. Note that in those settings the unit at which there are structural constraints is a sentence. Also note that there are many sentences available so that online training methods such as discriminative training of structured perceptrons can be used to learn structured predictors effectively in those settings. In contrast, for the cognates setting the unit at which there are structural constraints is the entire set of cognates for a language pair and there is only one such unit in existence (for a given language pair). We call this a single overarching global structure to make the distinction clear. The method we present in this paper deals with a single overarching global structure on the predictions of all instances in the {\em entire} problem space for a task. For this type of setting, there is only a {\em single} global structure in existence, contrasted with the situation of there being many sentences each imposing a global structure on the tagging decisions for that individual sentence. Hence, previous structured prediction methods that require numerous instances each having a structured output on which to train parameters via methods such as perceptron training are inapplicable to the cognates setting. In this paper we present methods for rescoring effectively in settings with a single overarching global structure and show their applicability to improving the performance of cognates detection. Still, we note that philosophically our  method builds on previous structured prediction methods since in both cases there is a similar intuition in that we're using higher-level structural properties to inform and accordingly alter our system's predictions of values for subitems within a structure.

In section~\ref{algorithm} we present our methods for performing rescoring of matrices based on global constraints such as those that apply for cognates detection. The key intuition behind our approach is that the scoring of word pairs for cognateness ought not be made independently as is currently done, but rather that global constraints ought to be taken into account to inform and potentially alter system scores for word pairs based on the scores of other word pairs. In section~\ref{experiments} we provide results of experiments testing the proposed methods on the cognates detection task on multiple datasets with multiple language pairs under multiple conditions. We show that the new methods complement and effectively improve performance over state of the art performance achieved by combining the major research breakthroughs that have taken place in cognates detection research to date.  Complete precision-recall curves are provided that show the full range of performance improvements over the current state of the art that are achieved. Summary measurements of performance improvements, depending on the language pair and dataset, range from 6.73 absolute MaxF1 percentage points to 16.75 absolute MaxF1 percentage points and from 5.58 absolute 11-point interpolated average precision percentage points to 17.19 absolute 11-point interpolated average precision percentage points.
Section~\ref{discussion} discusses the results and possible extensions of the method. Section~\ref{conclusions} wraps up with the main conclusions.

\section{Algorithm} \label{algorithm}

While our focus in this paper is on using global constraints to improve cognates detection, we believe that our method is useful more generally. We therefore abstract out some of the specifics of cognates detection and present our algorithm more generally in this section, with the hope that it will be able to be used in the future for other applications in addition to cognates detection. None of our abstraction harms understanding of our method's applicability to cognates detection and the fact that the method may be more widely beneficial does not in any way detract from the utility we show it has for improving cognates detection. 

A common setting is where one has a set $X = \{x_1,x_2,...,x_n\}$ and a set $Y=\{y_1,y_2,...,y_n\}$ where the task is to extract $(x,y)$ pairs such that $(x,y)$ are in some relation $R$. Here are examples: 
\begin{itemize}
\item $X$ might be a set of states and $Y$ might be a set of cities and the relation $R$ might be ``is the capital of"; 
\item $X$ might be a set of images and $Y$ might be a set of people's names and the relation $R$ might be ``is a picture of";
\item $X$ might be a set of English words and $Y$ might be a set of French words and the relation $R$ might be ``is cognate with".
\end{itemize}

A common way these problems are approached is that a model is trained that can score each pair $(x,y)$ and those pairs with scores above a threshold are extracted. We propose that often the relation will have a tendency, or a hard constraint, to satisfy particular properties and that this ought to be utilized to improve the quality of the extracted pairs. 

The approach we put forward is to re-score each $(x,y)$ pair by utilizing scores generated for other pairs and our knowledge of properties of the relation being extracted. In this paper, we present and evaluate methods for improving the scores of each $(x,y)$ pair for the case when the relation is known to be one-to-one and discuss extensions to other situations. 

The current approach is to generate a matrix of scores for each candidate pair as follows:
\begin{equation} \label{matrix}
Score_{X,Y} =  \begin{bmatrix}
         s_{x_{1},y_{1}} & \cdots & s_{x_{1},y_{n}} \\
         \vdots      & \ddots & \vdots      \\
         s_{x_{n},y_{1}} & \cdots & s_{x_{n},y_{n}}
        \end{bmatrix}.
\end{equation}  Then those pairs with scores above a threshold are predicted as being in the relation. We now describe methods for sharpening the scores in the matrix by utilizing the fact that there is an overarching global structure on the predictions.  

\subsection{Reverse Rank}

We know that if $(x_i,y_j) \in R$, then $(x_k,y_j) \notin R$  for $k \neq i$ when $R$ is 1-to-1. 
We define $reverse\_rank(x_i,y_j) = |\{ x_k \in X | s_{x_k,y_j} \ge s_{x_i,y_j} \}|$.
Intuitively, a high reverse rank means that there are lots of other elements of $X$ that score better to $y_j$ than $x_i$ does; this could be evidence that $(x_i,y_j)$ is not in $R$ and ought to have a lower score. Alternatively, if there are very few or no other elements of $X$ that score better to $y_j$ than $x_i$ does this could be evidence that $(x_i,y_j)$ is in $R$ and ought to have a higher score. In accord with this intuition, we use reverse rank as the basis for rescaling our scores as follows:
\begin{equation}
score_{RR}(x_i,y_j) = \frac{s_{x_{i},y_{j}}}{reverse\_rank(x_i,y_j)}.
\end{equation}

\subsection{Forward Rank}

Analogous to reverse rank, another basis we can use for adjusting scores is the forward rank. 
We define $forward\_rank(x_i,y_j) = |\{ y_k \in Y | s_{x_i,y_k} \ge s_{x_i,y_j} \}|$. We then scale the scores analogously to how we did with reverse ranks via an inverse linear function.\footnote{For both reverse rank and forward rank we also experimented with exponential decay and step functions, but found that simple division by the ranks worked as well or better than any of those more complicated methods.}

\subsection{Combining Reverse Rank and Forward Rank}

For combining reverse rank and forward rank, we present results of experiments doing it two ways. The first is a 1-step approach:
\begin{equation}
score_{RR\_FR\_1step}(x_i,y_j) = \frac{s_{x_{i},y_{j}}}{product},
\end{equation}

where \\
\begin{equation}
\begin{split}
product = & reverse\_rank(x_i,y_j) \times \\
& forward\_rank(x_i,y_j).
\end{split}
\end{equation}

The second combination method involves first computing the reverse rank and re-adjusting every score based on the reverse ranks. Then in a second step the new scores are used to compute forward ranks and then those scores are adjusted based on the forward ranks. We refer to this method as RR\_FR\_2step. 

\subsection{Maximum Assignment}

If one makes the assumption that all elements in $X$ and $Y$ are present and have their partner element in the other set present with no extra elements and the sets are not too large, then it is interesting to compute what the `maximal assignment' would be using the Hungarian Algorithm to optimize:

\begin{equation}
\begin{aligned}
& \underset{Z \in X \times Y}{\text{max}}
& & \sum_{(x,y) \in Z} score(x,y) \\
& \text{s.t.} & &  (x_i,y_j) \in Z \Rightarrow (x_k,y_j) \notin Z, \forall k \ne i \\
& & &  (x_i,y_j) \in Z \Rightarrow (x_i,y_k) \notin Z, \forall k \ne j. \\
\end{aligned}
\end{equation}

We do this on datasets where the assumptions hold and see how close our methods get to the Hungarian maximal assignment at similar points of the precision-recall curves. For our larger datasets where the assumptions don't hold, the Hungarian either can't complete due to limited computational resources or it functioned poorly in comparison with the performance of our reverse rank and forward rank combination methods. 

\section{Experiments} \label{experiments}

Our goal is to test whether using the global structure algorithms we described in section~\ref{algorithm} can significantly boost performance for cognates detection. To test this hypothesis, our first step is to implement a system that uses state of the art research results to generate the initial score matrices as a current state of the art system would currently do for this task. To that end, we implemented a baseline state of the art system that uses the information sources that previous research has found to be helpful for this task such as phonetic information, word context information, temporal context information, word frequency information, and word burstiness information \cite{kondrak2001,mann2001,schafer2002,klementiev2006,irvine2013}. Consistent with past work \cite{irvine2013}, we use supervised training to learn the weights for combining the various information sources. The system combines the sources of information by using weights learned by an SVM (Support Vector Machine) on a small seed training set of cognates\footnote{The small seed set was randomly selected and less than 20\% in all cases. It was not used for testing. Note that using this data to optimize performance of the baseline system makes our baseline even stronger and makes it even harder for our new rescoring method to achieve larger improvements.} to optimize performance. This baseline system obtains state of the art performance on cognates detection. Using this state of the art system as our baseline, we investigated how much we could improve performance beyond current state of the art levels by applying the rescoring algorithm we described in section~\ref{algorithm}. We performed experiments on three language pairs: French-English, German-English, and Spanish-English, with different text corpora used as training and test data. The different language pairs and datasets have different levels of performance in terms of their baseline current state of the art score matrices. In the next few subsections, we describe our experimental details.

\subsection{Lemmatization}

We used morphological analyzers to convert the words in text corpora to lemma form. For English, we used the NLTK WordNetLemmatizer \cite{bird2009}. For French, German, and Spanish we used the TreeTagger \cite{schmid1994}.

\subsection{Word Context Information} \label{wordContext}

We used the Google N-Gram corpus \cite{michel2010}.
For English we used the English 2012 Google 5-gram corpus, for French we used the French 2012 Google 5-gram corpus, for German we used the German 2012 Google 5-gram corpus, and for Spanish we used the Spanish 2012 Google 5-gram corpus. From these corpora we compute word context similarity scores across languages using Rapp's method \cite{rapp1995,rapp1999}.
The intuition behind this method is that cognates are more likely to occur in correlating context windows and this statistic inferred from large amounts of data captures this correlation. 

\subsection{Frequency Information}

The intuition is that over large amounts of data cognates should have similar relative frequencies. We compute our relative frequencies by using the same corpora mentioned in the previous subsection. 

\subsection{Temporal Information}

The intuition is that cognates will have similar temporal distributions \cite{klementiev2006}. To compute the temporal similarity we use newspaper data and convert it to simple daily word counts. For each word in the corpora the word counts create a time series vector. The Fourier transform is computed on the time series vectors. Spearman rank correlation is computed on the transform vectors. For English we used the English Gigaword Fifth Edition\footnote{Linguistic Data Consortium Catalog No. LDC2011T07}. For French we used French Gigaword Third Edition\footnote{Linguistic Data Consortium Catalog No. LDC2011T10}. For Spanish we used Spanish Gigaword First Edition\footnote{Linguistic Data Consortium Catalog No. LDC2006T12}.  The German news corpora were obtained by web crawling \url{http://www.tagesspiegel.de/} and extracting the news articles. 

\subsection{Word Burstiness}

The intuition is that cognates will have similar burstiness measures \cite{church1995}. For word burstiness we used the same corpora as for the temporal corpora. 

\subsection{Phonetic Information} \label{phonetic}

The intuition is that cognates will have correspondences in how they are pronounced. For this, we compute a measurement based on Normalized Edit Distance (NED). 

\subsection{Combining Information Sources}

We combine the information sources by using a linear Support Vector Machine to learn weights for each of the information sources from a small seed training set of cognates. 
So our final score assigned to a candidate cognate pair $(x,y)$ is:

\begin{equation}
score(x,y) = \sum_{m \in metrics}{w_mscore_m(x,y)},
\end{equation}
where $metrics$ is the set of measurements such as phonetic similarity measurements, word burstiness similarity, relative frequency similarity, etc. that were explained in subsections~\ref{wordContext} through \ref{phonetic}; $w_m$ is the learned weight for metric $m$; and $score_m(x,y)$ is the score assigned to the pair $(x,y)$ by metric $m$. 

The scores such assigned represent a state of the art approach for filling in the matrix identified in equation~\ref{matrix}. At this point the matrix of scores would be used to predict cognates. 
We now turn to evaluation of the use of the global constraint rescoring methods from section~\ref{algorithm} for improving performance beyond the state of the art levels.

\subsection{Using Global Constraints to Rescore} \label{rescoring}

\begin{figure}
\begin{center}
\centerline{\includegraphics[width=\columnwidth]{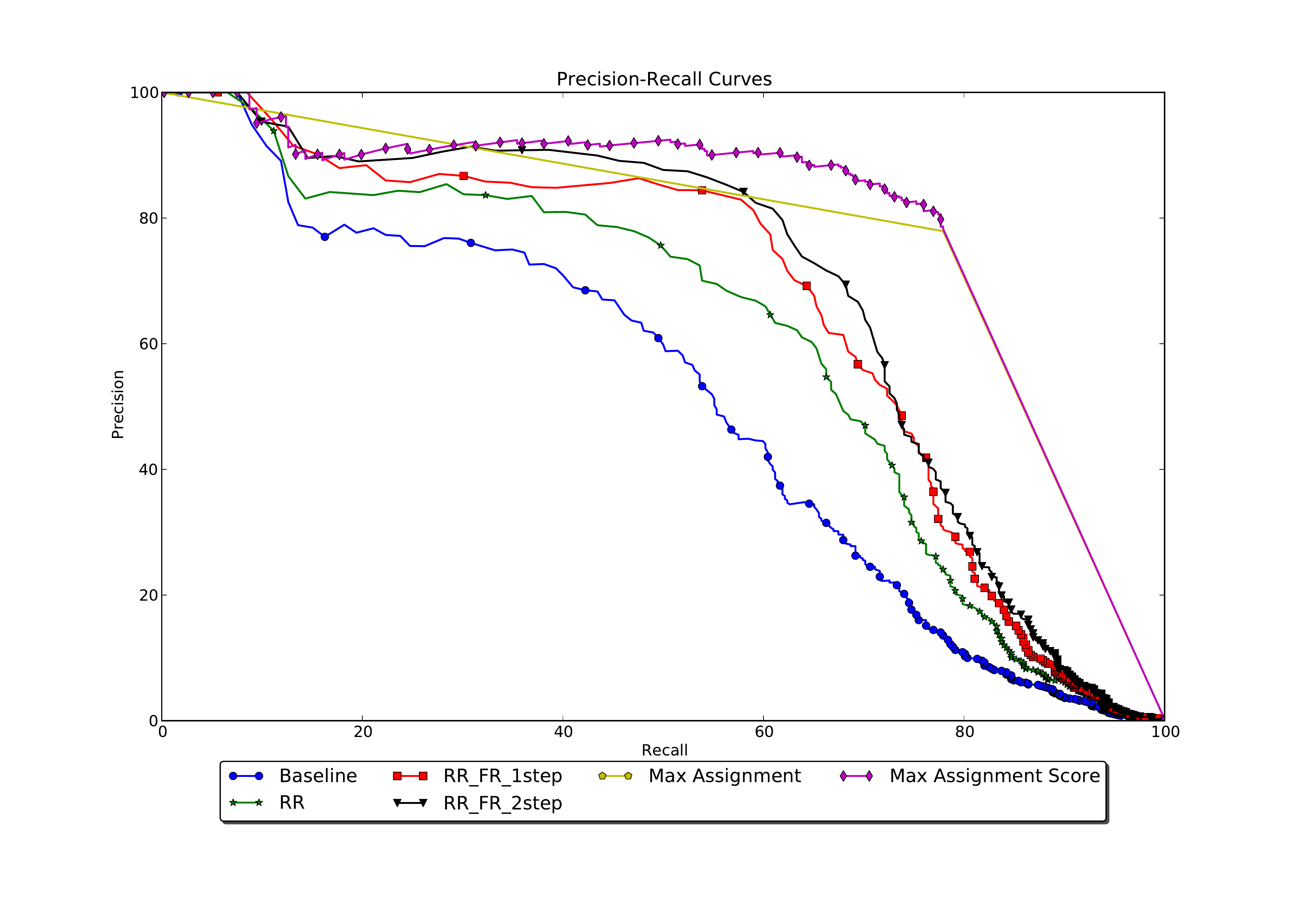}}
\caption{Precision-Recall Curves for French-English. Baseline denotes state of the art performance.}
\label{f:FrenchEnglish}
\end{center}
\end{figure}

\begin{figure}
\begin{center}
\centerline{\includegraphics[width=\columnwidth]{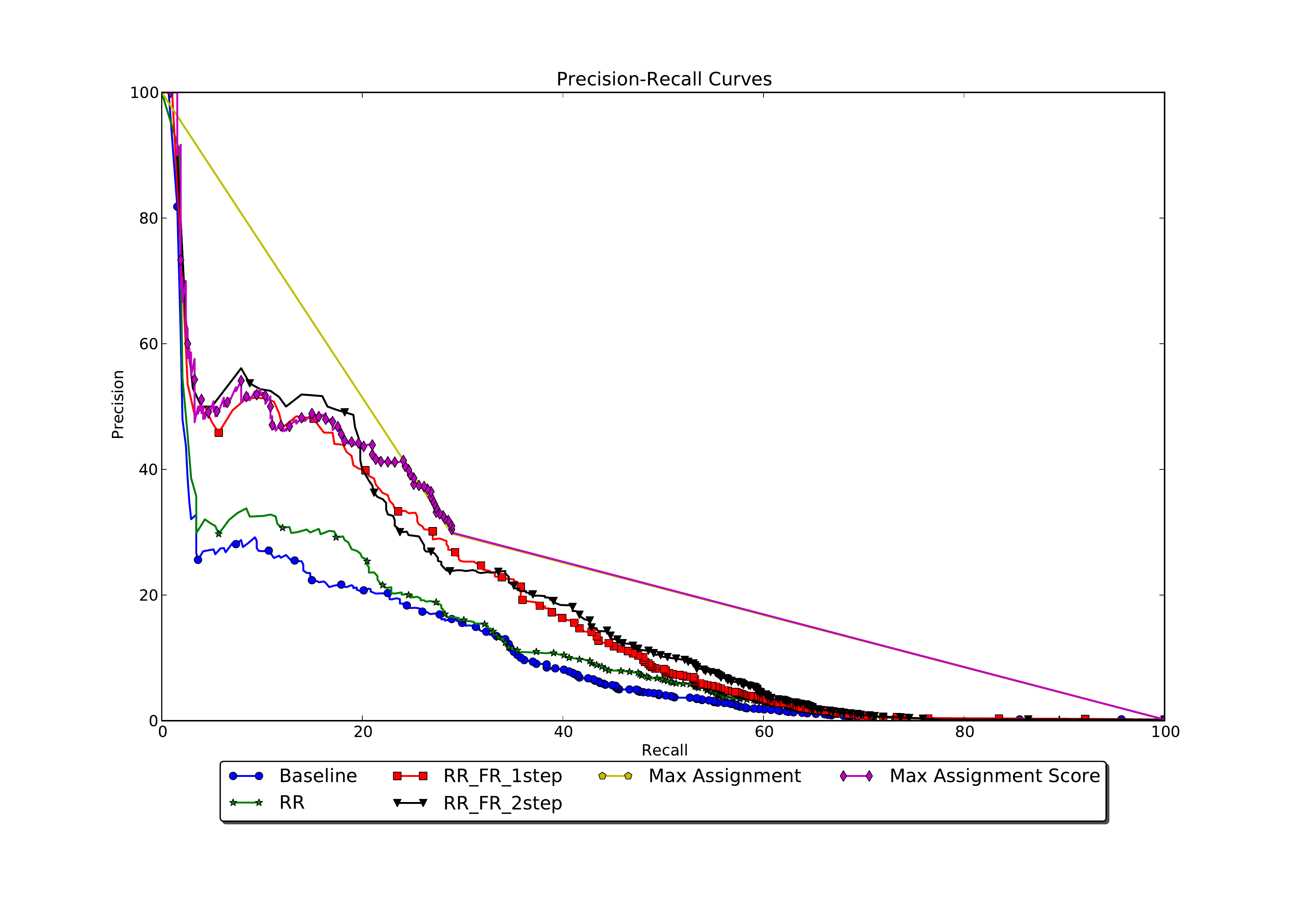}}
\caption{Precision-Recall Curves for German-English. Baseline denotes state of the art performance.}
\label{f:GermanEnglish}
\end{center}
\end{figure}

\begin{figure}[t]
\begin{center}
\centerline{\includegraphics[width=\columnwidth]{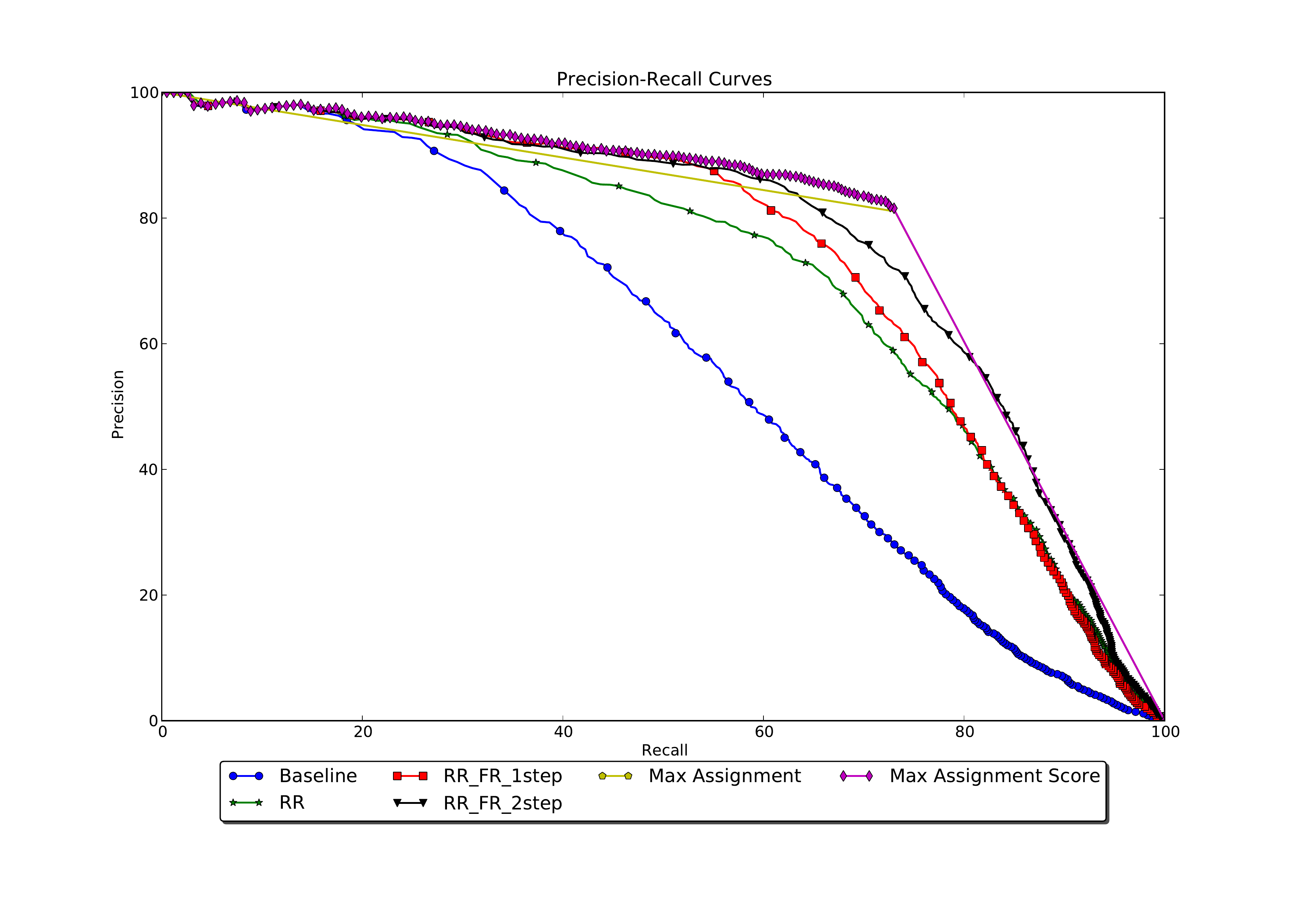}}
\caption{Precision-Recall Curves for Spanish-English. Baseline denotes state of the art performance.}
\label{f:SpanishEnglish}
\end{center}
\end{figure}

For our cognates data we used the French-English pairs from \cite{bergsma2007} and the German-English and Spanish-English pairs from \cite{beinborn2013}. 
Figure~\ref{f:FrenchEnglish} shows the precision-recall\footnote{Precision and recall are the standard measures used for systems that perform search. Precision is the percentage of predicted cognates that are indeed cognate. Recall is the percentage of cognates that are predicted as cognate. We vary the threshold that determines cognateness to generate all points along the Precision-Recall curve. We start with a very high threshold enabling precision of 100\% and lower the threshold until recall of 100\% is reached. In particular, we sort the test examples by score in descending order and then go down the list of scores in order to complete the entire precision-recall curve.} curves for French-English, Figure~\ref{f:GermanEnglish} shows the performance for German-English, and Figure~\ref{f:SpanishEnglish} shows the performance for Spanish-English. Note that state of the art performance (denoted in the figures as Baseline) has very different performance across the three datasets, but in all cases the systems from section~\ref{algorithm} that incorporate global constraints and perform rescoring greatly exceed current state of the art performance levels. The Max Assignment is really just the single point that the Hungarian finds. We drew lines connecting it, but keep in mind those lines are just connecting the single point to the endpoints. Max Assignment Score traces the precision-recall curve back from the Max Assignment by steadily increasing the threshold so that only points in the maximum assignment set with scores above the increasing threshold are predicted as cognate. 

For the non-max assignment curves, it is sometimes helpful to compute a single metric summarizing important aspects of the full curve. For this purpose, maxF1 and 11-point interpolated average precision are often used. MaxF1 is the F1 measure (i.e., harmonic mean of precision and recall) at the point on the precision-recall curve where F1 is highest. The interpolated precision $p_{interp}$ at a given recall level $r$ is defined as the highest precision level found for any recall level $r' \ge r$:
\begin{equation}
p_{interp}(r) = max_{r' \ge r} p(r').
\end{equation}
The 11-point interpolated average precision (11-point IAP) is then the average of the $p_{interp}$ at $r = 0.0, 0.1, ..., 1.0$.
Table~\ref{t:FrenchEnglish} shows these performance measures for French-English,
Table~\ref{t:GermanEnglish} shows the results for German-English, and Table~\ref{t:SpanishEnglish} show the results for Spanish-English. 
In all cases, using global structure greatly improves upon the state of the art baseline performance. In \cite{bergsma2007}, for French-English data a result of 66.5 11-point IAP is reported for a situation where word alignments from a bitext are available and a result of 77.7 11-point IAP is reported for a situation where translation pairs are available in large quantities. The setting considered in the current paper is much more challenging since it does not use bilingual dictionaries or word alignments from bitexts. The setting in the current paper is the one mentioned as future work on page 663 of \cite{bergsma2007}: "In particular, we plan to investigate approaches that do not require the bilingual dictionaries or bitexts to generate training data."

\begin{table}
\begin{center}
\begin{small}
\begin{sc}
\begin{tabular}{|c|c|c|}
\hline
Method & Max F1 & 11-point IAP \\ \hline
Baseline & 54.92 & 50.99 \\ \hline
RR & 62.94 & 59.62 \\ \hline
RR\_FR\_1step & 68.35 & 64.42 \\ \hline
RR\_FR\_2step & 69.72 & 67.29 \\ \hline
\end{tabular}
\end{sc}
\end{small}
\end{center}
\caption{French-English Performance. {\sc Baseline} indicates current state of the art performance.}
\label{t:FrenchEnglish}
\end{table}

\begin{table}
\begin{center}
\begin{small}
\begin{sc}
\begin{tabular}{|c|c|c|}
\hline
Method & Max F1 & 11-point IAP \\ \hline
Baseline & 21.38 & 16.25 \\ \hline
RR & 22.71 & 17.80 \\ \hline
RR\_FR\_1step & 28.68 & 22.37 \\ \hline
RR\_FR\_2step & 28.11 & 21.83 \\ \hline
\end{tabular}
\end{sc}
\end{small}
\end{center}
\caption{German-English Performance. {\sc Baseline} indicates current state of the art performance.}
\label{t:GermanEnglish}
\end{table}

\begin{table}
\begin{center}
\begin{small}
\begin{sc}
\begin{tabular}{|c|c|c|}
\hline
Method & Max F1 & 11-point IAP \\ \hline
Baseline & 56.26 & 57.03 \\ \hline
RR & 68.52 & 69.33 \\ \hline
RR\_FR\_1step & 70.66 & 71.47 \\ \hline
RR\_FR\_2step & 73.01 & 74.22 \\ \hline
\end{tabular}
\end{sc}
\end{small}
\end{center}
\caption{Spanish-English Performance. {\sc Baseline} indicates current state of the art performance.}
\label{t:SpanishEnglish}
\end{table}

Note that the evaluation thus far is a bit artificial for real cognates detection because in a real setting you wouldn't only be selecting matches for relatively small subsets of words that are guaranteed to have a cognate on the other side. Such was the case for our evaluation where the French-English set had approx. 600 cognate pairs, the German-English set had approx. 1000 pairs, and the Spanish-English set had approx. 3000 pairs. In a real setting, the system would have to consider words that don't have a cognate match in the other language and not only words that were hand-selected and guaranteed to have cognates. We are not aware of others evaluating according to this much more difficult condition, but we think it is important to consider especially given the potential impacts it could have on the global structure methods we've put forward. Therefore, we run a second set of evaluations where we take the ten thousand most common words in our corpora for each of our languages, which contain many of the cognates from the standard test sets and we add in any remaining words from the standard test sets that didn't make it into the top ten thousand. We then repeat each of the experiments under this much more challenging condition. With approx. ten thousand squared candidates, i.e., approx. 100 million candidates, to consider for cognateness, this is a large data condition. The Hungarian didn't run to completion on two of the datasets due to limited computational resources. On French-English it completed, but achieved poorer performance than any of the other methods. This makes sense as it is designed when there really is a bipartite matching to be found like in the artificial yet standard cognates evaluation that was just presented. When confronted with large amounts of words that create a much denser space and have no match at all on the other side the all or nothing assignments of the Hungarian are not ideal. The reverse rank and forward rank rescoring methods are still quite effective in improving performance although not by as much as they did in the small data results from above. 

\begin{figure}
\begin{center}
\centerline{\includegraphics[width=\columnwidth]{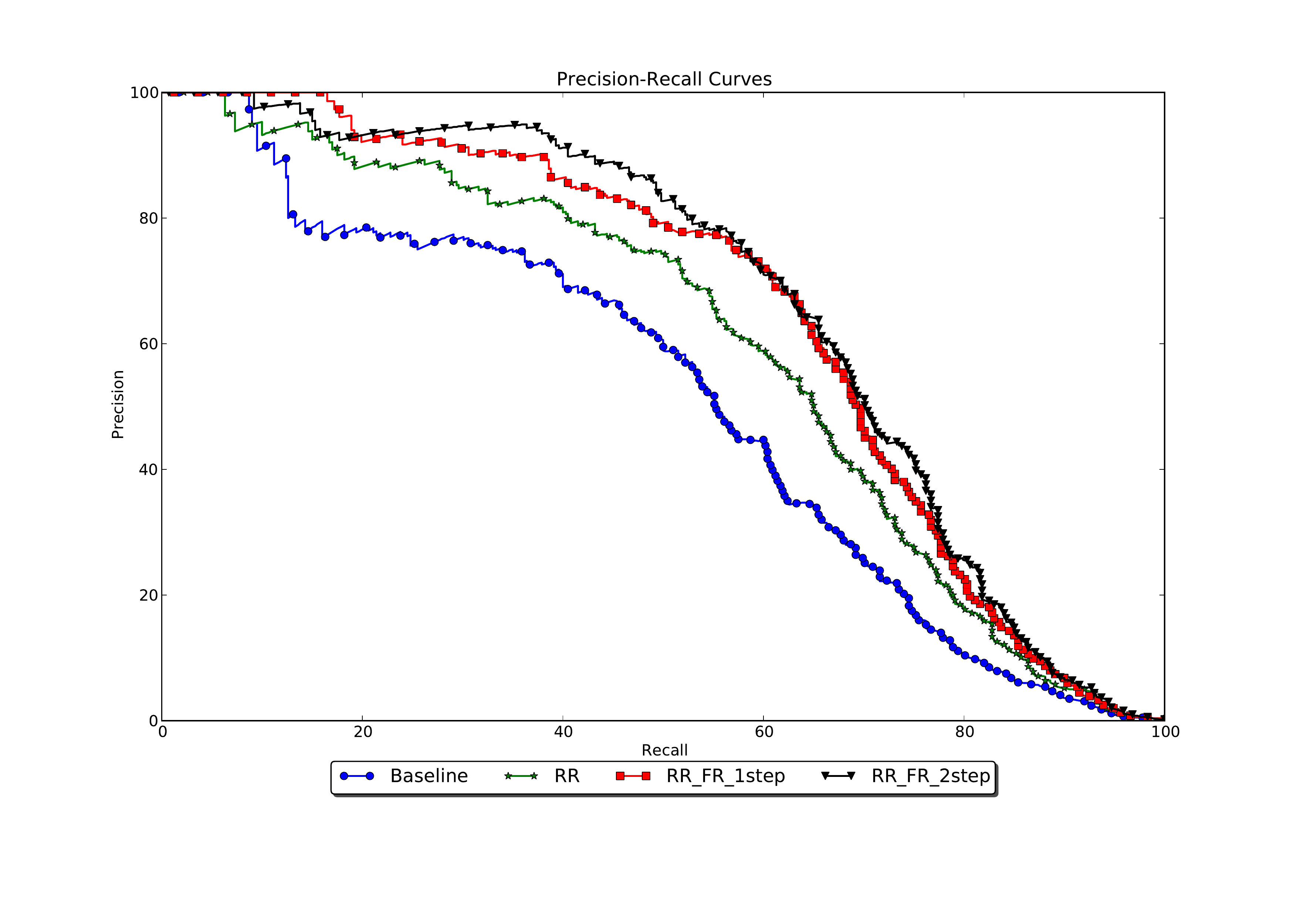}}
\caption{Precision-Recall Curves for French-English (large data). Note that Baseline denotes state of the art performance.}
\label{f:FrenchEnglishLarge}
\end{center}
\end{figure}

\begin{figure}[t]
\begin{center}
\centerline{\includegraphics[width=\columnwidth]{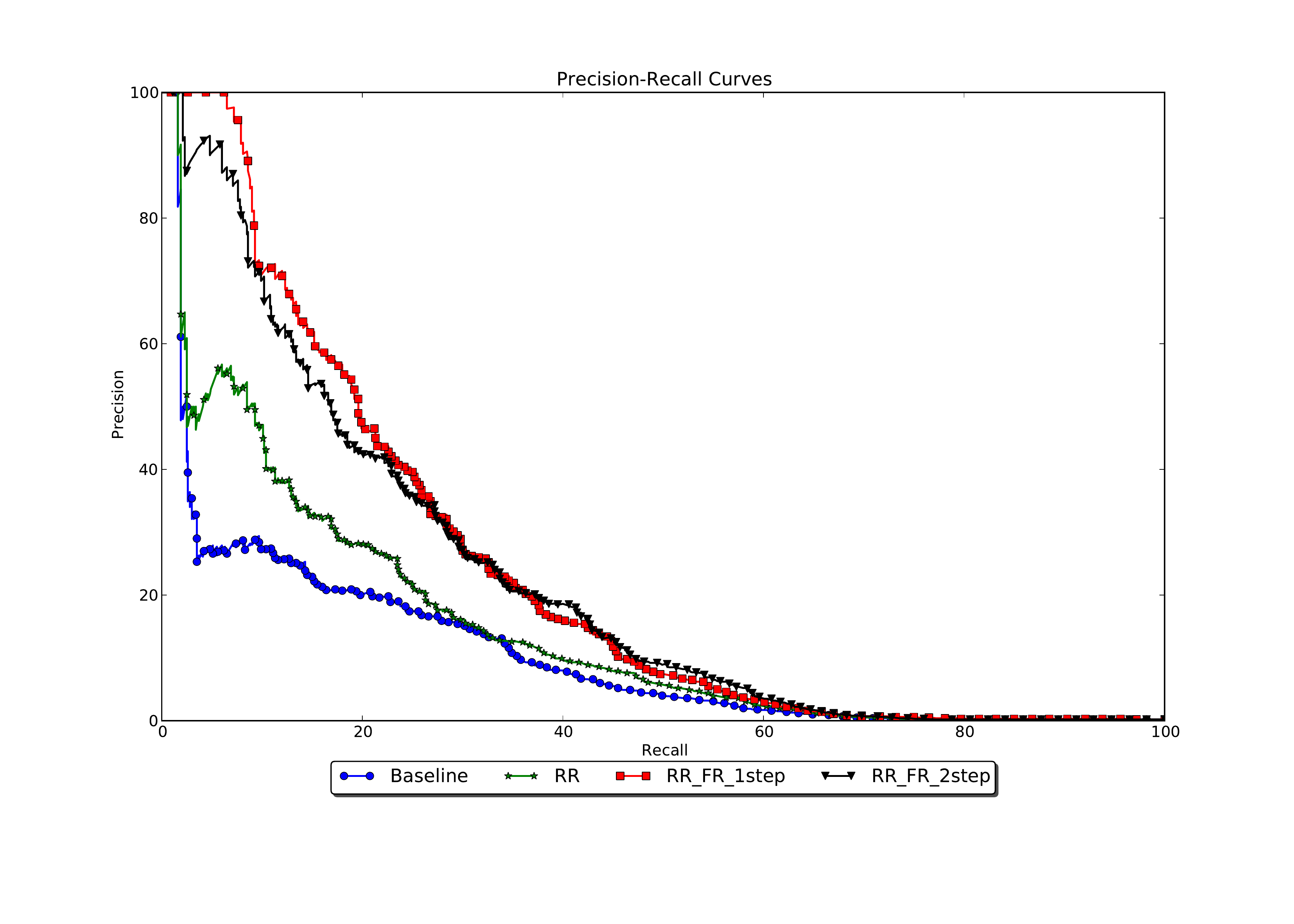}}
\caption{Precision-Recall Curves for German-English (large data). Note that Baseline denotes state of the art performance.}
\label{f:GermanEnglishLarge}
\end{center}
\end{figure}

\begin{figure}
\begin{center}
\centerline{\includegraphics[width=\columnwidth]{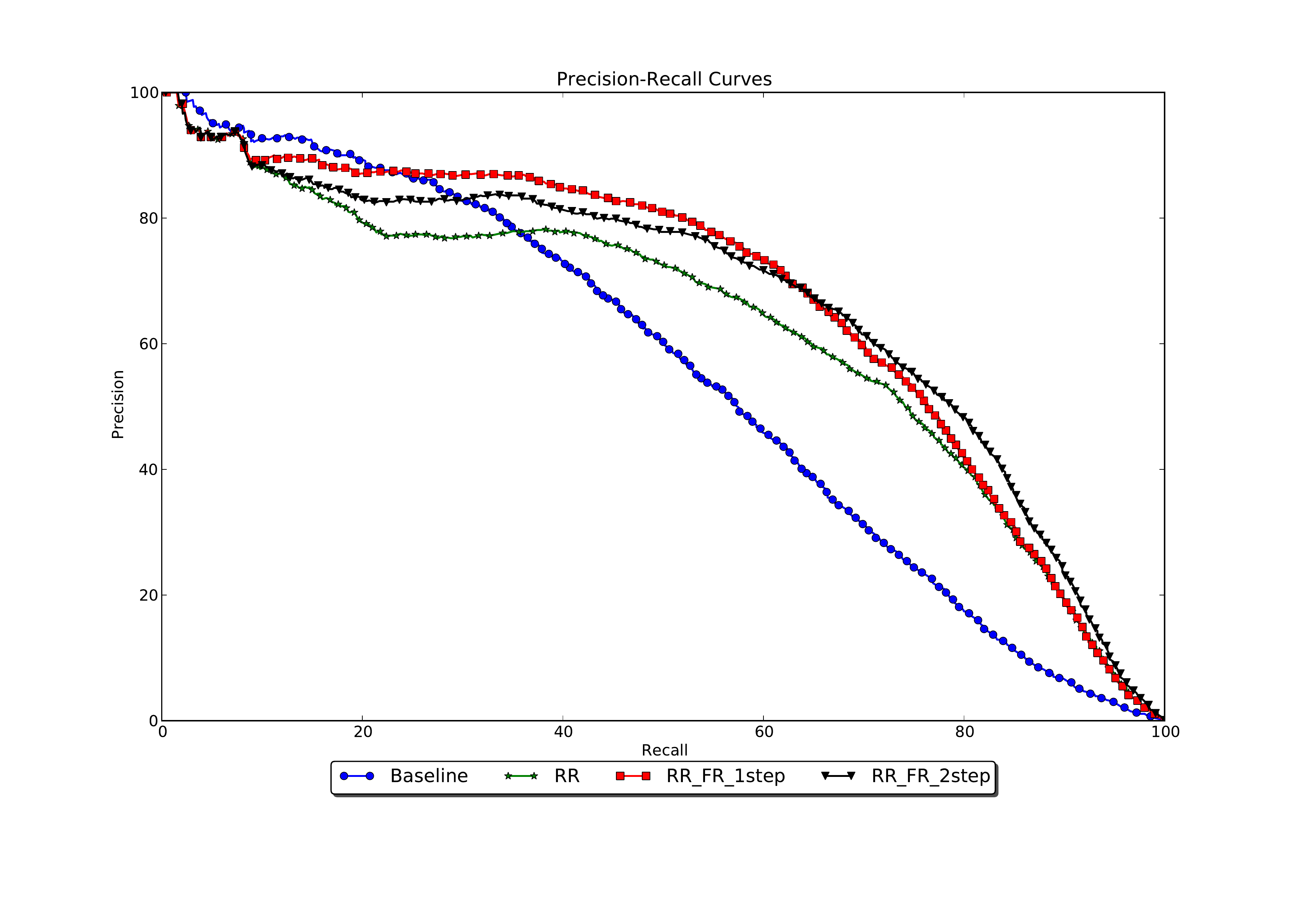}}
\caption{Precision-Recall Curves for Spanish-English (large data). Note that Baseline denotes state of the art performance.}
\label{f:SpanishEnglishLarge}
\end{center}
\end{figure}

Figure~\ref{f:FrenchEnglishLarge} shows the full precision-recall curves for French-English for the large data condition, Figure~\ref{f:GermanEnglishLarge} shows the curves for German-English for the large data condition, and Figure~\ref{f:SpanishEnglishLarge} shows the results for Spanish-English for the large data condition.   

Tables \ref{t:FrenchEnglishLarge} through \ref{t:SpanishEnglishLarge} show the summary metrics for the three language pairs for the large data experiments. We can see that the reverse rank and forward rank methods of taking into account the global structure of interactions among predictions is still helpful, providing large improvements in performance even in this challenging large data condition over strong state of the art baselines that make cognate predictions independently of each other and don't do any rescoring based on global constraints. 

\begin{table}
\begin{center}
\begin{small}
\begin{sc}
\begin{tabular}{|c|c|c|}
\hline
Method & Max F1 & 11-point IAP \\ \hline
Baseline & 55.08 & 51.35 \\ \hline
RR & 60.88 & 58.79 \\ \hline
RR\_FR\_1step & 65.87 & 63.55 \\ \hline
RR\_FR\_2step & 65.76 & 65.26 \\ \hline
\end{tabular}
\end{sc}
\end{small}
\end{center}
\caption{French-English Performance (large data). {\sc Baseline} indicates state of the art performance.}
\label{t:FrenchEnglishLarge}
\end{table}

\begin{table}
\begin{center}
\begin{small}
\begin{sc}
\begin{tabular}{|c|c|c|}
\hline
Method & Max F1 & 11-point IAP \\ \hline
Baseline & 21.25 & 16.17 \\ \hline
RR & 24.78 & 19.13 \\ \hline
RR\_FR\_1step & 30.72 & 24.97 \\ \hline
RR\_FR\_2step & 30.34 & 24.86 \\ \hline
\end{tabular}
\end{sc}
\end{small}
\end{center}
\caption{German-English Performance (large data). {\sc Baseline} indicates state of the art performance.}
\label{t:GermanEnglishLarge}
\end{table}

\begin{table}[!]
\begin{center}
\begin{small}
\begin{sc}
\begin{tabular}{|c|c|c|}
\hline
Method & Max F1 & 11-point IAP \\ \hline
Baseline & 54.75 & 54.55 \\ \hline
RR & 62.52 & 61.42 \\ \hline
RR\_FR\_1step & 66.45 & 65.89 \\ \hline
RR\_FR\_2step & 66.38 & 65.5 \\ \hline
\end{tabular}
\end{sc}
\end{small}
\caption{Spanish-English Performance (large data). {\sc Baseline} indicates state of the art performance.}
\label{t:SpanishEnglishLarge}
\end{center}
\end{table}

\section{Discussion} \label{discussion}

We believe that this work opens up new avenues for further exploration. A few of these include the following:
\begin{itemize}
\item investigating the utility of applying and extending the method to other applications such as Information Extraction applications, many of which have similar global constraints as cognates detection;
\item investigating how to handle other forms of global structure including tendencies that are not necessarily hard constraints;
\item developing more theory to more precisely understand some of the nuances of using global structure when it's applicable and making connections with other areas of machine learning such as semi-supervised learning, active learning, etc.; and
\item investigating how to have a machine {\em learn} that global structure exists and {\em learn} what form of global structure exists.
\end{itemize}

\section{Conclusions} \label{conclusions}

Cognates detection is an interesting and challenging task. Previous work has yielded state of the art approaches that create a matrix of scores for all word pairs based on optimized weighted combinations of component scores computed on the basis of various helpful sources of information such as phonetic information, word context information, temporal context information, word frequency information, and word burstiness information. However, when assigning a score to a word pair, the current state of the art methods do not take into account scores assigned to other word pairs. We proposed a method for rescoring the matrix that current state of the art methods produce by taking into account the scores assigned to other word pairs. The methods presented in this paper are complementary to existing state of the art methods, easy to implement, computationally efficient, and practically effective in improving performance by large amounts. Experimental results reveal that the new methods significantly improve state of the art performance in multiple cognates detection experiments conducted on standard freely and publicly available datasets with different language pairs and various conditions such as different levels of baseline performance and different data size conditions, including with more realistic large data size conditions than have been evaluated with in the past.

\bibliography{paper}
\bibliographystyle{acl_natbib}

\end{document}